# Benchmarking and Evaluation of AI Models in Biology: Outcomes and Recommendations from the CZI Virtual Cells Workshop

Elizabeth Fahsbender [1], Alma Andersson [2], Jeremy Ash [3], Polina Binder [4], Daniel Burkhardt [4], Benjamin Chang [5], Georg K. Gerber [6,7,8,9], Anthony Gitter [10,11], Patrick Godau [12,13,14], Ankit Gupta [15], Genevieve Haliburton [1], Siyu He [16], Trey Ideker [17,18,19], Ivana Jelic [20], Aly Khan [21,22,23,24], Yang-Joon Kim [25], Aditi Krishnapriyan [26], Jon M. Laurent [27], Tianyu Liu [28], Emma Lundberg [15,29,30], Shalin B. Mehta [25], Rob Moccia [31], Angela Oliveira Pisco [32], Katherine S. Pollard [33,34,35], Suresh Ramani [4], Julio Saez-Rodriguez [36,37,38], Yasin Senbabaoglu [25], Elana Simon [16], Srinivasan Sivanandan [32], Gustavo Stolovitzky [39], Marc Valer [20], Bo Wang [29,41], Xikun Zhang [16], James Zou [16,42], Katrina Kalantar [1]

[1] Chan Zuckerberg Initiative
[2] Genentech Inc
[3] Johnson & Johnson Innovative Medicine
[4] NVIDIA
[5] Sanger Institute
[6] Harvard Medical School
[7] Division of Computational Pathology, Bringham and Women's Hospital
[8] Massachusetts Host-Microbiome Center
[9] Harvard-MIT Health Sciences & Technology
[10] Department of Biostatistics and Medical Informatics, University of Wisconsin-Madison
[11] Morgridge Institute for Research
[12] German Cancer Research Center (DKFZ)
[13] National Center for Tumor Diseases (NCT)
[14] Faculty of Mathematics and Computer Science, Heidelberg University
[15] Department of Protein Science, Science for Life Laboratory, KTH Royal Institute of Technology
[16] Department of Biomedical Data Science, Stanford
[17] Division of Human Genomics and Precision Medicine, Department of Medicine, University of California San Diego
[18] Department of Bioengineering, University of California San Diego
[19] Department of Computer Science and Engineering, University of California San Diego
[20] Formerly Chan Zuckerberg Initiative
[21] Departments of Pathology and Family Medicine, University of Chicago
[22] Toyota Technical Institute at Chicago
[23] Institute for Population and Precision Health, University of Chicago
[24] Chan Zuckerberg Biohub Chicago
[25] Chan Zuckerberg, Biohub San Francisco
[26] University of California, Berkeley
[27] FutureHouse
[28] Yale University
[29] Department of Bioengineering, Stanford University
[30] Department of Pathology, Stanford University
[31] Valid, Inc.
[32] Insitro, Inc
[33] Institute for Human Genetics, University of California, San Francisco
[34] Gladstone Institutes
[35] Department of Epidemiology and Biostatistics University of California San Francisco
[36] European Molecular Biology Laboratory, European Bioinformatics Institute (EMBL-EBI)
[37] Faculty of Medicine Heidelberg University
[38] Institute for Computational Biomedicine Heidelberg University Hospital
[39] Department of Pathology at NYU Grossman School of Medicine
[40] Genetics and Genomic Sciences, Icahn School of Medicine at Mount Sinai
[41] Department of Developmental Biology, Stanford University
[42] Electrical Engineering; and Computer Science, Stanford University

## Summary

Artificial intelligence holds immense promise for transforming biology, yet a lack of standardized, cross-domain, benchmarks undermines our ability to build robust, trustworthy models. Here, we present insights from a recent workshop that convened machine learning and computational biology experts across imaging, transcriptomics, proteomics, and genomics to tackle this gap. We identify major technical and systemic bottlenecks—such as data heterogeneity and noise, reproducibility challenges, biases, and the fragmented ecosystem of publicly available resources—and propose a set of recommendations for building benchmarking frameworks that can efficiently compare AI/ML models of biological systems across tasks and data modalities. By promoting high-quality data curation, standardized tooling, comprehensive evaluation metrics, and open, collaborative platforms, we aim to accelerate the development of robust benchmarks for AI-driven Virtual Cells. These benchmarks are crucial for ensuring rigor, reproducibility, and biological relevance, and will ultimately advance the field toward integrated models that drive new discoveries, therapeutic insights, and a deeper understanding of cellular systems.

## Introduction

The development of AI-driven Virtual Cells represents one of the most ambitious frontiers in biological modeling. These computational representations of cellular function aim to integrate multi-scale, multi-modal data to simulate the behavior of molecules, cells, and tissues across diverse biological states [1]. Such advancements promise to accelerate discoveries, guide experimental studies, and enable high-fidelity simulations that will transform biomedical research and therapeutic development. However, realizing this vision requires reliable, standardized, and biologically meaningful evaluation methods to assess AI models. Benchmarking plays a crucial role in ensuring that these models are not only powerful but also trustworthy, reproducible, and biologically relevant. Benchmarks are also critical for driving progress; when diverse approaches consistently plateau at a certain performance level, this can indicate a fundamental limitation, suggesting the need for more or new types of data to drive further progress. This creates a crucial feedback loop where data informs model development, models are evaluated through benchmarks, and the results of these benchmarks, in turn, can highlight areas where further data generation is essential. Without rigorous benchmarking, the pursuit of Virtual Cells risks being driven by models that excel in narrow, artificial settings but fail when applied to complex biological problems.

Artificial intelligence (AI) is rapidly transforming the landscape of biological research, offering unprecedented opportunities for accelerating discovery across diverse fields. We have now seen how well-crafted benchmarks and their supporting ecosystems have driven landmark innovations, as was the case with the CASP challenge in driving the advances seen for protein structure prediction by

AlphaFold [2,3]. There are numerous examples of success tackling various aspects within the space of benchmarking for biological AI [4–12]. However, as we move beyond single domains and consider the technical foundations necessary to achieve integrated modeling approaches in service of Virtual Cell models [1,13–17], new challenges emerge while existing ones are compounded. Surveying efforts across domains reveals that the lack of standardized benchmarks and evaluation metrics for AI models in biology remains a significant barrier to realizing the full potential of these powerful tools. Without consistent standards, it becomes difficult to compare the performance of different models, hindering reproducibility, and ultimately slowing progress in the field [18,19].

Recognizing the urgent need for community-driven efforts to address this challenge, a recent workshop convened by the Chan Zuckerberg Initiative (CZI) brought together a diverse group of experts across imaging, proteomics, and genomics, from academia, industry, and non-profit open science organizations. The workshop fostered a collaborative environment to identify challenges, solutions and opportunities in existing benchmarking approaches. The recommendations of this group are summarized here.

Building on the vision for building AI Virtual Cells [1], this effort emphasizes the need for benchmarking infrastructure tailored to the unique requirements of large-scale models, ensuring broad applicability across diverse domains in biology. By gathering input from a wide range of stakeholders, the initiative aims to outline a benchmarking system that benefits researchers across fields and accelerates the development and adoption of valuable AI tools for biological discovery.

## Challenges in Benchmarking for AI in Biology

**Data**

In contrast to other machine learning fields with large, purpose-built datasets containing millions of observations, datasets for biological problems tend to be smaller, heterogeneous, imbalanced, and noisy. These limitations arise from the high costs of data generation, the challenges of data acquisition, and the complexity of curation. Biological data is often particularly affected by data sparsity and batch effects, arising from technical, experimental, and biological sources including differences in instruments, reagents, and calibration, as well as variations in environmental conditions, sample handling, and processing protocols [20–22]. These issues can obscure true biological signals and hinder accurate data analysis and interpretation [23].

Furthermore, curation of appropriate metadata and ground truth labels can be both laborious and subject to the same technical challenges. Even in cases where robust labels may be obtained, lack of standardization in data formats, schemas, and ontologies reduce the ability to share and compare data across different studies and platforms. Privacy concerns, especially when there is personally identifiable information (PII) or personal genomic information (PGI), further complicate data sharing. There is a tension between the need for open data sharing to foster collaboration and the importance of protecting sensitive patient information. Finally, in the context of AI systems, data leakage between training and evaluation datasets becomes a key concern. Given the challenges outlined above, it is often difficult to trace which data has been used for training purposes versus evaluation, thereby potentially inadvertently leading to inflated performance estimates.

**Reproducibility**

Reproducibility in benchmarking faces several challenges. Primarily, the complexity of code design, often involving diverse tools and implementations, makes debugging and maintenance difficult. This complexity can be compounded by a lack of clear, accessible documentation detailing the code and experimental setup. Without such documentation, understanding, replicating, and validating results becomes a significant hurdle. Furthermore, the absence of incentives to prioritize reproducibility discourages the creation and ongoing maintenance of well-documented and easily reproducible workflows. Although technical solutions exist, the current academic environment, which prioritizes speed of publication and novelty of findings, often fails to adequately reward the substantial effort required to create and maintain well-documented and easily reproducible workflows, resulting in inconsistent application of existing technical solutions.

**Biological relevance of evaluation metrics**

Unlike many machine learning tasks with well-defined objectives and quantifiable outcomes, biological hypotheses are often multifaceted and require approaches that go beyond a single performance metric. Focusing solely on single target measures can lead to distorted model development (Goodhart's law) and misallocation of research resources [24].

Within the context of biology, testing hypotheses frequently involves a series of experiments that span diverse modalities and methodologies, from in vitro to animal models to human clinical studies. This generates a wide range of data types, including text, imaging, and tabular data, each with special characteristics like compositionality, non-linear measurement noise, and sparse or missing values. Evaluating whether experimental data supports or refutes a hypothesis is a complex process involving quantitative and qualitative interpretation, consideration of prior scientific evidence, and peer consensus. Biological context is critical in these evaluations. For example, cellular behavior can vary significantly due to internal factors like genetic variants and external factors like signals from neighboring cells. While biologists have well-established processes for evaluating findings, these processes are not easily translated into standard machine learning benchmarks. This difficulty arises because it's challenging to translate complex biological questions into clear, quantifiable metrics that accurately reflect model performance in the specific biological context.

**Fragmented Ecosystem**

Benchmarking efforts suffer from a fragmented ecosystem where data, leaderboards, models, publications, and resources are scattered across various platforms and repositories. This lack of a centralized location hinders collaboration and slows progress, by making it difficult to discover relevant benchmarks or compare results across different studies.

**Biases**

Unbiased data sets are critical to training and evaluating machine learning models, however, bias in biological datasets can arise from multiple sources. In addition to technical and measurement biases, as discussed in the **Data** section, systematic biases within the scientific process, study design, and recruitment phases can confound the evaluation and comparison of models [25,26]. Selection of which experiments researchers choose to pursue can be influenced by cost, study complexity, or previous work, potentially leading to datasets that don't comprehensively cover all relevant biological or chemical spaces and therefore limit model generalizability. For example, known genetic ancestral and gender bias pervades large data repositories in the biomedical domain [27,28]. Furthermore, studies have demonstrated that biomedical research disproportionately focuses on well-characterized genes or those already prominent in the literature, leading to significant gaps in understanding functional diversity and creating a feedback loop that neglects unexplored regions of the genome [29,30]. Compounding the issue, the significant underreporting of studies with null or negative results leads to publication bias that skews the scientific literature toward positive findings [31,32];[32].

### Benchmark Maintenance and Evolution

The rapid evolution of both biological technologies and AI models necessitates a dynamic approach to benchmarking. Evaluation metrics and datasets must be continuously re-evaluated and adapted to keep pace with advances. It is crucial to consider technological flux when designing a successful benchmarking pipeline. Novel methods may yield results that challenge existing evaluation criteria, while the emergence of new metrics may offer more nuanced insights into performance. Failing to account for this simultaneous evolution can lead to misleading comparisons and hinder progress.

### Community Development

Developing and adopting standardized benchmarks demands a collaborative effort, uniting the diverse expertise and resources of pharma, biotech, non-profits, and academia. However, differing incentive structures, the interdisciplinary nature of benchmarking, and the lack of a shared language across these diverse domains can hinder community development. Pharmaceutical companies may prioritize benchmarks aligned with their commercial interests, while academic researchers might focus on more theoretical aspects. Bridging these gaps requires fostering a shared understanding of the value and purpose of standardized benchmarks. Despite these challenges, broad community participation is crucial to ensure inclusivity, transparency, and widespread adoption of best practices. By bringing together diverse perspectives and expertise across machine learning researchers, computational biologists, and experimentalists, the community can create robust, relevant, and impactful benchmarks that drive innovation and improve research quality across the field.

## Recommendations for addressing these challenges

**Invest in high-quality data generation and curation** prioritizing investments in annotated datasets designed specifically for benchmarking purposes. Curation efforts should seek to align to existing ontologies to facilitate interoperability [33]. To prevent data leakage upon release of these datasets, they should be released with clear guidelines or hosted in a way that makes them only accessible for testing [18,34].

**Develop standardized tooling that ensures benchmarks are robust and reproducible**. To enhance reproducibility, focus should be placed on improved code management through modular design, standardized documentation practices that clearly detail the codebase and experimental setup, and establishing incentive structures that recognize and reward efforts towards following open source best-practices and creating reproducible benchmarks. In some cases, workforce development initiatives may further improve the ubiquity of these practices [35]. With respect to benchmarking, these needs should be addressed across the three main components of reproducibility outlined by McDermott et al [19]: *technical replicability* – made possible by sharing of reproducible, versioned, code i.e. via containerization approaches similar to those employed by OpenProblems [5], *statistical replicability* – made possible by effective data splitting and resampling processes similar to those employed by Polaris [12], and lastly, *conceptual replicability* – made possible via documentation of workflows, annotation of datasets, and inclusion of supporting metadata.

**Utilize a multi-faceted evaluation approach, incorporating diverse metrics that capture different aspects of model performance**, including accuracy, robustness, and generalizability. Metrics should be selected through close collaboration between a variety of domain experts (e.g. cell biologists, physicians, machine learning researchers, and biostatisticians) and impacted stakeholders (e.g. companies, FDA) - who can help align metrics with real-world biological applications and interpretation needs, thereby further strengthening trust in the evaluation process [24]. See *Ensure that benchmarks are relevant across diverse research questions* below. Additionally, selected metrics should be accompanied by clear articulation of metric selection rationale, transparent documentation of evaluation procedures, and careful validation of the metrics themselves [36]. Metric implementations and any aggregation need to be robust for variations in models or data, such as missing values, sample inter-dependencies or algorithmic non-determinism [37]. While multiple complementary metrics are necessary to fully characterize model performance, their aggregation and presentation should remain interpretable, providing flexible insights into both model successes as well as failure modes. This can be achieved through organizing evaluations into distinct subtasks, employing interactive visualizations that reveal performance complexities [38] and breaking down performance characteristics across different data distributions, such as data from different laboratories or time points [20].

**Ensure that benchmarks are relevant across diverse research questions.** There is a need to involve domain experts to ensure the biological relevance of selected tasks, evaluation criteria, and results interpretation. To engage experts, it will be important to communicate in the language of the domain, potentially providing support to translate experimental language into formulation of evaluation metrics. See also, *Engage a community of interdisciplinary experts*, below.

**Create a centralized platform to serve as a hub for sharing benchmarks, datasets, models, evaluation tools and best practices**. A centralized platform and development of common standard formats that facilitate exchange could foster a more collaborative environment, facilitate knowledge transfer, and accelerate advancements in the field. Such a resource should enable ease of access to

benchmarks and supporting assets that are implemented in line with the recommendations set forth by the community. Where centralization is not feasible (due to restrictions in data, model IP, or otherwise), federated systems with cross-referencing via standardized formats may provide another option.

**Engage a community of interdisciplinary experts to create best practices**. The community of experts who can contribute to benchmarking AI models in biology should be defined broadly and inclusively: computational modelers; software, data, and infrastructure engineers; data scientists, biologists and clinicians. These groups each play distinct but critical functions to the success of a benchmarking initiative. Community engagement can arise naturally as well as through supported mechanisms – for example, via a cross-disciplinary committee charged with providing structured guidance towards the evolution of benchmarks over time. Forums for open discussion of progress should be maintained to ensure community feedback can be incorporated from researchers in each domain on an ongoing basis to ensure the right benchmarks, data generation, and evaluation strategies are employed. Drawing inspiration from the success of the Critical Assessment of Structure Prediction (CASP) experiments [39], recurring opportunities to discuss progress on hard scientific problems and regular, transparent reporting of that progress represent important aspects of engaging the community [40,41].

**Ensure benchmarks remain relevant and updated as the field progresses**. A flexible and forward-looking approach to benchmarking is essential to ensure that evaluations remain relevant and informative in this era of rapid technological change. There should be a clear mechanism and willingness to deprecate benchmarks that no longer accurately reflect the current state of the field or the capabilities of emerging technologies. These decisions should be guided by community discussions, as outlined above.

**Foster collaboration and knowledge sharing across sectors to promote shared resources to accelerate benchmarking efforts**. Best-practices arise through collaborative discussion and scientists with the relevant expertise come from many different backgrounds and geographic locations. Traditionally, community building is accomplished with in-person meetings, but to be inclusive of the international community, distributed, asynchronous, and local options must be considered.

# Conclusion

Benchmarking is foundational for building credible, trustworthy, and impactful AI models for biology. Through this workshop, the community identified critical challenges – from data heterogeneity and lack of reproducibility to fragmented ecosystems and misaligned incentives – and outlined concrete recommendations to address them.

To realize the vision of AI Virtual Cells, we must foster a cultural shift: prioritizing rigorous evaluation as a top-line goal, investing in high-quality data generation and curation, adopting standardized tools prioritizing multi-faceted evaluation metrics, and building interoperable, community-driven platforms for sharing resources and knowledge. Underlying the success of these aims are the sustained, interdisciplinary collaborations that bridge the diverse expertise of experimentalists, computer scientists, clinicians, and engineers. This path requires collective effort, transparency, and continual iteration. By embracing these principles, we can build robust benchmarking ecosystems to accelerate innovation while ensuring that AI models remain aligned with the complex biological realities they aim to represent. With community commitment, we can turn the vision of Virtual Cells into a scientific reality, transforming our ability to understand, diagnose, and treat human disease.